\documentclass[letterpaper, 10 pt, conference]{ieeeconf}

\IEEEoverridecommandlockouts
\usepackage{amsmath}
\usepackage{amssymb}
\usepackage{booktabs}
\usepackage{graphicx}
\usepackage[caption=false,font=footnotesize]{subfig}
\usepackage{makecell}
\usepackage{multirow}
\usepackage{tabularx}
\usepackage[table]{xcolor}
\usepackage{cite}
\usepackage{url}
\usepackage[hidelinks]{hyperref}
\usepackage{placeins}
\usepackage{float}

\newcolumntype{L}[1]{>{\raggedright\arraybackslash}p{#1}}
\newcolumntype{R}[1]{>{\raggedleft\arraybackslash}p{#1}}
\newcolumntype{C}[1]{>{\centering\arraybackslash}p{#1}}
\title{\LARGE \bf
Joint-Health-Aware Reinforcement Learning for Robust Robot Control Under Actuator Degradation
}

\author{Haifa Zhang\qquad Yijing Wang\qquad Zhiqiang Zuo\\[0.4em]
\normalsize School of Electrical and Information Engineering\\
\normalsize Tianjin University, Tianjin, China}

\hypersetup{pdftitle={Joint-Health-Aware Reinforcement Learning for Robust Robot Control Under Actuator Degradation},pdfauthor={Haifa Zhang, Yijing Wang, Zhiqiang Zuo},pdfsubject={Research preprint}}
\begin{document}

\maketitle
\thispagestyle{plain}
\pagestyle{plain}

\begin{abstract}
Robots deployed in unstructured environments are susceptible to actuator-authority degradation, which reduces the effective control authority of individual actuators and can cause reinforcement learning policies trained under nominal dynamics to fail. Existing robustness approaches rely on discrete fault labels, broad domain randomization, or global latent variables, often without explicitly modeling continuous per-joint residual actuation capability. We propose a joint-health-aware reinforcement learning framework for robust robot control under simulated actuator degradation. Our framework encodes actuator condition in a continuous joint-health vector and models degradation as multiplicative attenuation of executed actions. A dedicated HealthNet estimator infers per-joint health from recent observations and actions, allowing the policy to redistribute control effort across healthy and degraded joints. We evaluate the method on Walker, Go2, Cassie, and Panda-based ManiSkill manipulation tasks under diverse degradation masks and severity levels. The method outperforms nominal and degradation-trained no-health baselines on the main evaluated metrics. Relative to the nominal baseline, task-level locomotion returns improve by up to 250.1\% and manipulation success rates by up to 15.2 percentage points; in the high-degradation Cassie group, return increases from 3.94 to 40.50 (+928.0\%). Ablations support the contribution of per-joint health estimation, action-degradation modeling, and health-conditioned control.
\end{abstract}

\section{Introduction}
\begin{figure}[!t]
\centering
\subfloat[Walker: 3 degraded joints\label{fig:same-damage-walker}]{%
  \includegraphics[width=0.95\columnwidth]{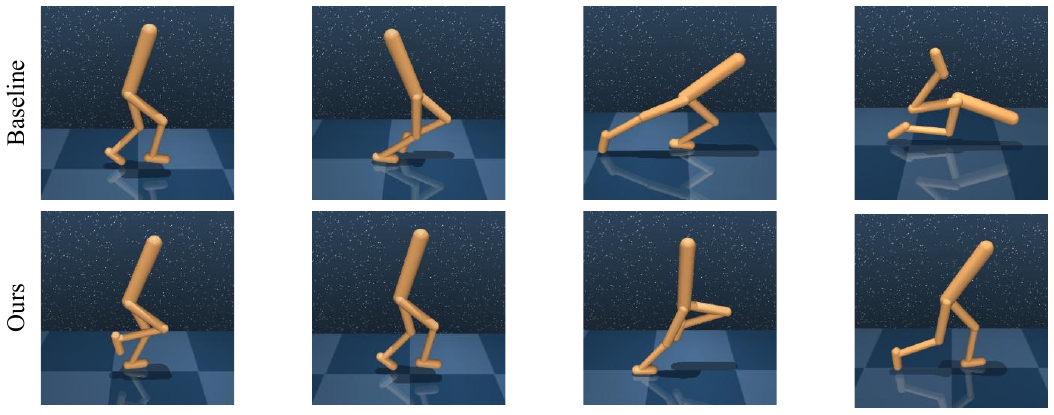}}
\par\vspace{-0.2em}
\subfloat[Go2: 5 degraded joints\label{fig:same-damage-go2}]{%
  \includegraphics[width=0.95\columnwidth]{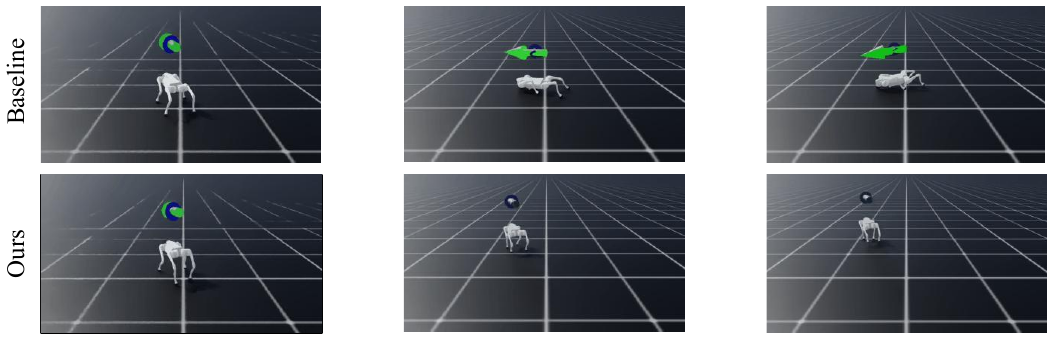}}
\par\vspace{-0.2em}
\subfloat[Cassie: 8 degraded joints\label{fig:same-damage-cassie}]{%
  \includegraphics[width=0.95\columnwidth]{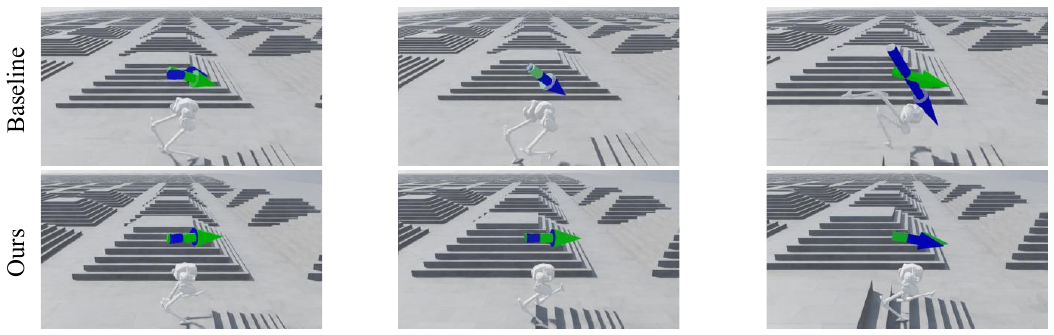}}
\par\vspace{-0.2em}
\subfloat[ManiSkill: 3 degraded joints\label{fig:same-damage-maniskill}]{%
  \includegraphics[width=0.95\columnwidth]{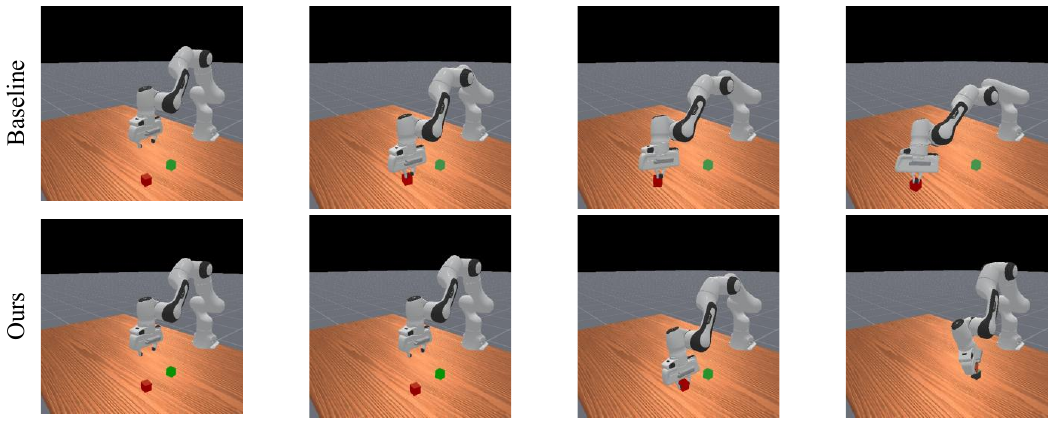}}
\caption{Same joint degradation, different response: the baseline degrades, while the health-conditioned policy adapts.}
\label{fig:same-damage-adaptation}
\vspace{-1.0em}
\end{figure}

Robots deployed in unstructured environments may experience actuator degradation or partial loss of control authority. Such failures are fundamentally different from ordinary sensing noise or mild dynamics perturbations. When a joint degrades, the command issued by the policy may no longer be faithfully realized by the corresponding actuator: a weakened hip, knee, calf, or wrist can still receive large control commands, but the resulting motion may be attenuated or nearly absent. This joint-level action--response mismatch is especially harmful for locomotion and manipulation, where successful behavior depends on coordinated multi-joint control. Fig.~\ref{fig:same-damage-adaptation} illustrates how a nominal controller and a health-conditioned policy respond differently to the same degradation on each platform.

While existing approaches enhance robustness to dynamics shifts, few explicitly represent how much actuation capability remains at each individual joint. Fault detection and diagnosis techniques typically describe failures using discrete labels, binary indicators, or predefined fault categories~\cite{khalastchi2018fddrobots,zhang2008ftcReview}. Domain randomization improves average robustness by training policies over diverse dynamics, but the controller is not directly informed about which actuator is degraded~\cite{tobin2017domain,peng2018dynamicsrandomization,rajeswaran2017epopt}. Online adaptation and latent-context schemes infer hidden dynamics from recent interaction history, yet the inferred variables are usually global or unstructured, lacking direct correspondence to individual joints~\cite{yu2017uposi,kumar2021rma,hafner2019dreamer}. As a result, these schemes provide limited explicit information for redistributing control effort under heterogeneous partial actuator degradation.

Achieving robust control under actuator-authority degradation motivates an explicit representation of residual per-joint actuation capability. A degraded robot cannot simply replay its original motion pattern, nor should it abandon partially impaired joints entirely. Rather, the controller needs to estimate which joints remain functional and how severely they are degraded, then use that information to redistribute control effort. This motivates coupling per-joint health estimation with policy conditioning, so that estimated residual capability directly informs action generation.

To this end, we propose a joint-health-aware reinforcement learning framework for robust robot control under actuator-authority degradation. The framework characterizes actuator condition via a continuous joint-health vector \(h_t \in [0,1]^J\), where each element captures the residual control authority of one actuator. During training, sampled health vectors modulate executed actions through a multiplicative degradation gate, producing structured action--response mismatch under simulated actuator degradation. A dedicated HealthNet estimator predicts current joint health from a short observation--action history, and the policy conditions on this prediction to adapt its actions. During deployment, true health is unavailable; the policy relies exclusively on estimated health to reduce dependence on degraded joints and exploit the remaining capability of healthier ones.

We evaluate the framework on Walker, Unitree Go2, Cassie, and Panda-based ManiSkill manipulation tasks under diverse degradation masks and severity levels. The method improves performance over nominal and degradation-trained no-health baselines. Relative to the nominal baseline, task-level locomotion returns improve by up to 250.1\% and manipulation success rates by up to 15.2 percentage points.

The main contributions of this work are threefold:
\begin{itemize}
    \item Actuator-authority degradation is formulated as a continuous per-joint health representation, and its effect is modeled through an action degradation mechanism, enabling finer-grained characterization of residual control authority than binary or categorical fault indicators.

    \item A joint-health-aware reinforcement learning framework with HealthNet is developed to estimate per-joint health from recent observation--action history and condition the control policy on the predicted health vector for degradation-aware control redistribution.

    \item The proposed method is evaluated across locomotion and manipulation benchmarks, including Walker, Unitree Go2, Cassie, and Panda-based ManiSkill tasks, demonstrating consistent robustness improvements under diverse joint-degradation patterns and severity levels.
\end{itemize}

\section{Related Work and Motivation}

\textbf{Robot learning algorithms and benchmark platforms.}
Deep reinforcement learning has become a standard tool for simulated and sim-to-real robot control, supported by policy-gradient and model-based algorithms such as PPO, TD-MPC, and TD-MPC2~\cite{schulman2017ppo,hansen2022tdmpc,hansen2024tdmpc2}. In locomotion, prior work has trained policies in simulation and transferred them to legged robots through actuator modeling, domain randomization, energy-aware rewards, imitation, and compact proprioceptive observations~\cite{heess2017emergence,haarnoja2018learning,tan2018sim,hwangbo2019learning,peng2020learningagile,rudin2022learning,li2021robustbiped,lee2020challengingterrain,miki2022perceptive,margolis2022walk,fu2021minimizing,fu2022coupling}. Typical benchmark platforms include MuJoCo/DMControl, Isaac Gym/Orbit/Isaac Lab, and ManiSkill environments~\cite{todorov2012mujoco,tassa2018dmcontrol,makoviychuk2021isaacgym,mittal2023orbit,mittal2025isaaclab,mu2021maniskill,gu2023maniskill2}.

\textbf{Robust reinforcement learning and online adaptation.}
Robust reinforcement learning and dynamics randomization have been widely used to improve robot control under dynamics shifts and deployment uncertainty. By training policies over randomized physical parameters, disturbances, terrains, or actuation variations, these methods enhance average robustness and facilitate sim-to-real transfer~\cite{jakobi1995radical,tobin2017domain,peng2018dynamicsrandomization,rajeswaran2017epopt,chebotar2019simopt,openai2018dexterous,akkaya2019rubik}. A parallel line of work infers hidden context from recent interaction history. Online system identification, meta-reinforcement learning, RMA, and DreamWaQ demonstrate that proprioceptive and action--observation histories can encode useful information about terrain, payload, friction, morphology changes, and other latent dynamics factors~\cite{yu2017uposi,nagabandi2019learning,kumar2021rma,nahrendra2023dreamwaq,finn2017maml,duan2016rl2,wang2016learning}. However, such methods typically infer global or unstructured context variables. Consequently, while they adapt to changing dynamics, they lack explicit estimates of residual actuation capability at each individual joint.

\textbf{Fault-tolerant control under joint degradation and faults.}
Learning-based fault-tolerant control has advanced considerably in damaged robot locomotion. Early biologically inspired work showed that robots can recover from damage by discovering compensatory behaviors~\cite{cully2015robots}. More recent progress trains policies directly under simulated joint failures, such as joint locking, power loss, masked actions, or adversarially selected damage cases~\cite{liu2022saving,yang2020faultaware}. Random Joint Masking trains quadrupedal policies with randomly disabled joints and a joint-state estimator, enabling locomotion under impaired-joint conditions~\cite{kim2024impaired}. Multi-task fault-tolerant controllers learn distinct behaviors for healthy motion, joint power loss, and joint locking within a hierarchical framework~\cite{hou2024multitask}. DreamFLEX estimates joint fault information and modulates the locomotion policy under rough-terrain anomaly conditions~\cite{lee2025dreamflex}, while AcL distills multiple fault-specific locomotion behaviors into a student policy for quadrupeds with powered-off joints~\cite{xu2025acl}.

These findings indicate that robots can continue moving after actuation faults, but most of them focus on fault adaptation rather than continuous joint-health modeling. Specifically, their fault representations are often binary, categorical, or tied to predefined failure modes such as locked joints, powered-off actuators, or masked action channels. Such representations can localize faults, yet provide little information about how severely each joint is degraded or whether it still has usable residual capability. In practice, actuator degradation is often gradual: a joint may continue to generate motion, but with reduced torque, degraded tracking, slower response, or intermittent effectiveness. Treating such a joint as either healthy or failed can overlook usable residual control authority.

\textbf{Joint-health-aware control.}
Our framework addresses this gap with a continuous joint-health vector whose elements quantify residual actuator authority. HealthNet estimates this vector from recent observation--action history and supplies it as explicit policy context. This representation captures heterogeneous partial degradation and supports control redistribution while retaining usable actuation capability. We evaluate this principle across Walker, Go2, Cassie, and Panda-based manipulation tasks, extending the evaluation beyond quadrupedal locomotion.

\section{Problem Formulation}

This work addresses robot control under actuator degradation. The nominal system is modeled as a Markov decision process \((\mathcal{S}, \mathcal{A}, P, r, \gamma)\), where \(s_t \in \mathcal{S}\) is the state, \(a_t \in \mathcal{A}\) is the policy command, \(P\) is the transition dynamics, \(r\) is the reward, and \(\gamma\) is the discount factor. The policy receives an observation \(o_t\), which may contain proprioceptive states, task information, and previous control signals depending on the environment.

At each instant \(t\), actuator degradation is represented by a joint health vector
\[
    h_t \in [0,1]^J ,
\]
where \(J\) is the number of controllable joints and \(h_{t,j}\) denotes the residual control authority of the \(j\)-th actuator. A value \(h_{t,j}=1\) corresponds to a fully functional joint, while smaller values indicate partial degradation; \(h_{t,j}\approx 0\) corresponds to a nearly disabled actuator. When degradation is fixed within an episode, \(h_t\) remains constant; we nevertheless use time-indexed notation to accommodate degradation injected during rollouts.

Degradation affects the action actually applied to the robot. For torque-control environments, we model the executed action as
\[
    a_t^e = h_t \odot a_t ,
\]
where \(\odot\) denotes element-wise multiplication. The degraded transition is therefore governed by
\[
    s_{t+1} \sim P(s_t, a_t^e).
\]
This formulation captures the key effect of actuator degradation: the commanded action and the executed action no longer match when a joint loses control authority.

During deployment, the true health vector is not directly observable. The agent must infer the current actuator condition from a recent window of interaction history and act accordingly. To this end, the history window is defined as
\[
    \xi_t = \{(o_{t-K}, a_{t-K}), \ldots, (o_{t-1}, a_{t-1}), o_t\}.
\]
The objective is to learn a health estimator and a control policy that maximize expected return under unknown joint health:
\[
    \max_{\theta,\phi} \;
    \mathbb{E}\left[\sum_{t=0}^{T} \gamma^t r_t \right],
    \quad
    \hat h_t = f_\theta(\xi_t), \quad
    a_t \sim \pi_\phi(o_t,\hat h_t).
\]
During simulation training, sampled \(h_t\) values generate degraded action execution and provide supervision for the health estimator. During deployment, only the predicted health \(\hat h_t\) is available to the policy.

\section{Methodology}

\begin{figure*}[!t]
\centering
\includegraphics[width=0.98\textwidth]{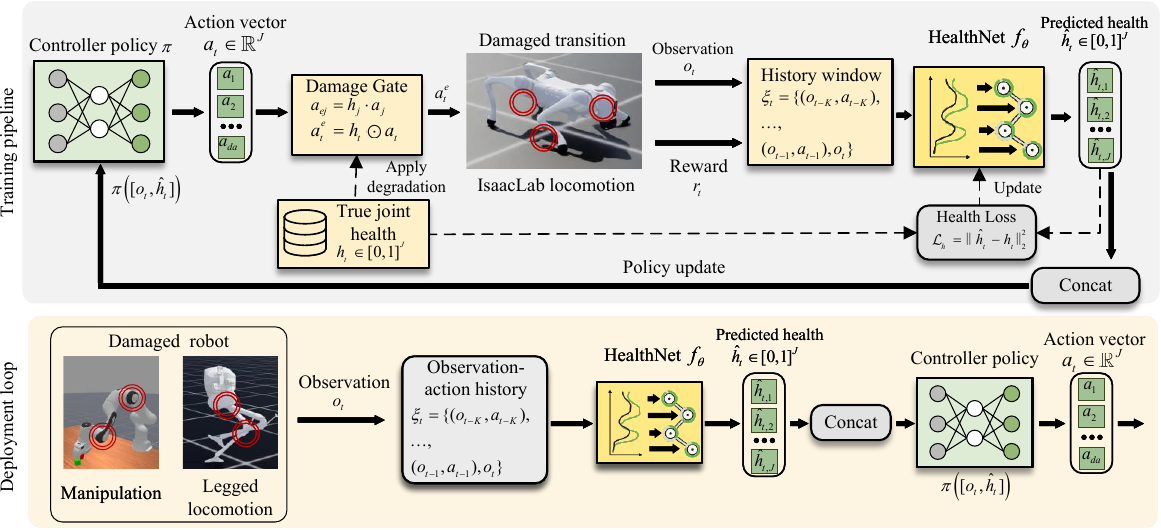}
\caption{Overview of joint-health-aware control. HealthNet estimates \(\hat h_t\) from recent action--response history, and the controller conditions on \((o_t,\hat h_t)\) under actuator degradation. During training, sampled true health \(h_t\) degrades the executed action and supervises HealthNet; during deployment, only the predicted health \(\hat h_t\) is used by the controller.}
\label{fig:overview}
\end{figure*}

The proposed scheme incorporates reinforcement learning with an explicit estimate of the robot's current joint-level actuation capability.
As shown in Fig.~\ref{fig:overview}, the framework consists of three components:
an action degradation model, a HealthNet estimator, and a health-conditioned control policy.
At each time step, HealthNet estimates the current joint health \(\hat h_t\) from recent observation--action history.
The policy then conditions on \((o_t,\hat h_t)\) to generate an action \(a_t\).
During simulation training, the true sampled health \(h_t\) degrades the executed action and provides supervision for HealthNet.
During deployment, \(h_t\) is unavailable and the policy uses only the predicted health \(\hat h_t\).

\subsection{Action Degradation Model}
\label{subsec:action-degradation}

Actuator degradation is modeled as an attenuation of effective control authority at each joint.
For a robot with \(J\) controllable joints, the current joint health is represented by
\(h_t \in [0,1]^J\), where \(h_{t,j}=1\) indicates a fully functional actuator and smaller values signify reduced actuation capability.
Each action dimension maps to a single controlled joint, so the action vector satisfies \(a_t \in \mathbb{R}^J\).

\begin{figure}[!t]
\centering
\includegraphics[width=\columnwidth]{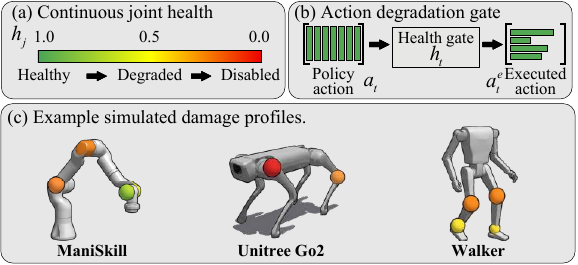}
\caption{Continuous joint-health representation and action degradation. The health vector \(h_t\) specifies residual joint capability, and the action degradation gate maps policy actions \(a_t\) to executed actions \(a_t^e\) under simulated actuator degradation.}
\label{fig:action-degradation}
\end{figure}

The executed action is obtained via a multiplicative health gate:
\begin{equation}
    a_t^e = h_t \odot a_t,
    \qquad
    a_{t,j}^e = h_{t,j} a_{t,j},
    \quad j=1,\ldots,J ,
    \label{eq:action_degradation}
\end{equation}
This formulation aligns with loss-of-actuator-effectiveness models in fault-tolerant control~\cite{zhang2002partialActuatorFailures,zhang2008ftcReview}.
In torque-control environments, \(a_t^e\) is directly applied as the degraded control command.
For target-position controllers, the policy action is interpreted as an offset from a default pose, and the degraded target is calculated as
\begin{equation}
    q_t^\star = q_0 + s a_t^e
    = q_0 + s (h_t \odot a_t),
    \label{eq:position_degradation}
\end{equation}
where \(q_0\) is the default joint configuration and \(s\) is the action scale.
Fig.~\ref{fig:action-degradation} illustrates this degradation process.

This representation covers fully healthy, partially weakened, and nearly disabled joints within a unified framework.
Unlike binary masking, a degraded joint is not always removed from control.
For example, a joint with \(h_{t,j}=0.6\) retains partial authority and can contribute useful motion; the policy should learn to compensate for its reduced authority through other joints.

\subsection{HealthNet: Per-Joint Health Estimation}
\label{subsec:healthnet}

The true health vector \(h_t\) is not available during deployment.
Therefore, the agent must infer the current actuator condition from recent interaction history.
The history window is defined as
\begin{equation}
    \xi_t =
    \{(o_{t-K},a_{t-K}),\ldots,(o_{t-1},a_{t-1}),o_t\}.
    \label{eq:history}
\end{equation}
This history includes past actions and observations together with the current observation. It excludes the current action \(a_t\), which is generated only after health estimation is complete.

\begin{figure}[!t]
\centering
\includegraphics[width=\columnwidth]{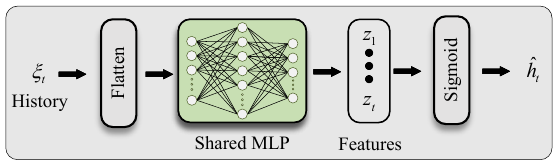}
\caption{HealthNet architecture. A recent action--response history is flattened and encoded by a shared multilayer perceptron, followed by feature projection and a sigmoid output for the predicted joint-health vector.}
\label{fig:healthnet}
\end{figure}

The proposed HealthNet estimates per-joint health as
\begin{equation}
    \hat h_t = f_\theta(\xi_t),
    \qquad
    \hat h_t \in [0,1]^J .
    \label{eq:healthnet}
\end{equation}
In the implementation, \(\xi_t\) is flattened and encoded by a shared multilayer perceptron, followed by a projection layer and a sigmoid output head:
\begin{equation}
    z_t = g_\theta(\xi_t),
    \qquad
    \hat h_t = \sigma(W_h z_t + b_h).
    \label{eq:healthnet_arch}
\end{equation}
This matches the HealthNet architecture in Fig.~\ref{fig:healthnet}.

The estimator is trained with an auxiliary health prediction loss:
\begin{equation}
    \mathcal L_h =
    \mathbb{E}_t
    \left[
    \frac{1}{J}
    \left\|
    \hat h_t - h_t
    \right\|_2^2
    \right].
    \label{eq:health_loss}
\end{equation}
The target \(h_t\) is available only during simulation training because it is sampled by the degradation generator.
In the deployment stage, no ground-truth health vector is assumed.

\begin{table*}[!t]
\centering
\caption{Main results under joint degradation. Locomotion gives mean episode return; manipulation gives final success rate. Gains are computed relative to the nominal baseline.}
\label{tab:main_results}
\small
\setlength{\tabcolsep}{6pt}
\begin{tabular*}{0.96\textwidth}{@{\extracolsep{\fill}}cccc>{\bfseries}cc@{}}
\toprule
\textbf{Domain} & \textbf{Task} & \textbf{Nominal} & \makecell{\textbf{Degradation-trained}\\\textbf{No-Health}} & \textbf{Ours} & \(\boldsymbol{\Delta}_{\text{\bfseries Nominal}}\) \\
\midrule
DMControl & Walker Walk       & 633.46 & 642.37 & 775.99 & +22.5\% \\
DMControl & Walker Run        & 237.24 & 266.70 & 370.14 & +56.0\% \\
DMControl & Walker WalkBack   & 489.24 & 588.58 & 736.08 & +50.5\% \\
DMControl & Walker RunBack    & 184.20 & 239.38 & 317.52 & +72.4\% \\
Isaac Lab & Go2 Flat          & 12.37  & 18.22  & 36.07  & +191.6\% \\
Isaac Lab & Go2 Rough         & 18.01  & 20.46  & 27.88  & +54.8\% \\
Isaac Lab & Cassie Flat       & 14.89  & 16.86  & 52.14  & +250.1\% \\
Isaac Lab & Cassie Rough      & 16.94  & 20.57  & 38.29  & +126.0\% \\
ManiSkill & StackCube         & 12.2\% & 11.1\%  & 27.4\% & +15.2 pp \\
ManiSkill & OpenCabinetDrawer & 59.6\% & 64.8\%  & 72.6\% & +13.0 pp \\
ManiSkill & PickSingleYCB     & 40.6\% & 42.0\%  & 46.5\% & +5.9 pp \\
\bottomrule
\end{tabular*}
\end{table*}

\subsection{Health-Conditioned Policy}
\label{subsec:health-conditioned-policy}

The predicted health vector serves as an explicit control context.
The control policy is parameterized as
\begin{equation}
    a_t \sim \pi_\phi(\cdot \mid o_t,\hat h_t),
    \label{eq:policy}
\end{equation}
or equivalently \(a_t=\pi_\phi(o_t,\hat h_t)\) for deterministic notation.
By conditioning on \(\hat h_t\), the policy can learn to reduce reliance on severely degraded joints and redistribute control effort toward joints with higher estimated health.

The health-conditioned input is formed by concatenating the observation and the predicted health vector:
\begin{equation}
    x_t = [o_t,\hat h_t].
    \label{eq:policy_input}
\end{equation}
The actor receives \(x_t\) to generate actions, and the critic employs the same health-conditioned context in an actor--critic implementation.

\subsection{Training Objective}

The policy and HealthNet are trained from the same degraded rollouts.
At each training step, the policy generates \(a_t\) using the predicted health \(\hat h_t\), while the environment executes the degraded action \(a_t^e\) according to the sampled true health \(h_t\).
The resulting transition and reward are used for reinforcement learning, and \(h_t\) also serves to supervise HealthNet.

The overall training loss is
\begin{equation}
    \mathcal L =
    \mathcal L_{\mathrm{actor}}
    + \lambda_v \mathcal L_{\mathrm{value}}
    + \lambda_h \mathcal L_h ,
    \label{eq:training_loss}
\end{equation}
where \(\mathcal L_{\mathrm{actor}}\) and \(\mathcal L_{\mathrm{value}}\) are the standard policy and value losses of the underlying reinforcement learning algorithm, and \(\mathcal L_h\) is the \emph{health prediction loss} in~\eqref{eq:health_loss}.
The coefficient \(\lambda_h\) governs the strength of health supervision.

During deployment, the degradation generator and health supervision are removed.
The robot observes \(o_t\), constructs \(\xi_t\), predicts \(\hat h_t=f_\theta(\xi_t)\), and executes the health-conditioned policy \(a_t\sim\pi_\phi(\cdot\mid o_t,\hat h_t)\).
Thus, the true joint health is not required at test time.

\section{Experiments}

\begin{figure*}[!t]
\centering
\subfloat[Rollout snapshots\label{fig:go2_case_rollout}]{%
  \includegraphics[width=0.92\textwidth]{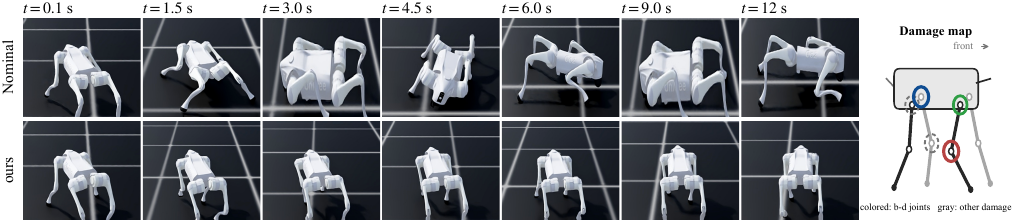}}
\vspace{-0.9em}

\subfloat[Health estimation\label{fig:go2_case_health}]{%
  \includegraphics[width=0.44\textwidth]{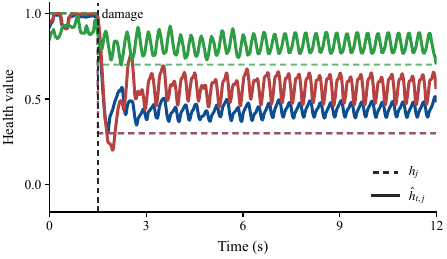}}
\hfill
\subfloat[Executed actions\label{fig:go2_case_executed_action}]{%
  \includegraphics[width=0.44\textwidth]{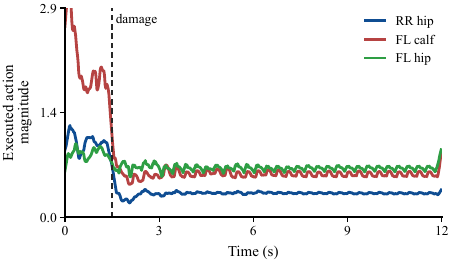}}
\vspace{-0.9em}

\subfloat[Raw policy actions\label{fig:go2_case_raw_action}]{%
  \includegraphics[width=0.44\textwidth]{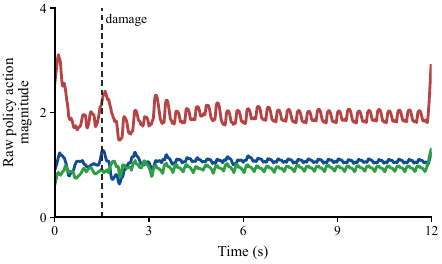}}
\hfill
\subfloat[Cumulative reward\label{fig:go2_case_reward}]{%
  \includegraphics[width=0.44\textwidth]{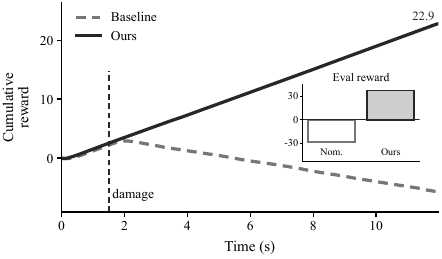}}
\caption{\textbf{Fixed-degradation Go2 case study.}
\textbf{a,} Paired rollout snapshots compare the nominal Go2 Flat baseline with the health-conditioned policy under five-joint degradation injected at \(t=1.5\,\mathrm{s}\).
\textbf{b,} True and estimated joint health for three representative degraded joints.
\textbf{c,d,} Executed and raw action magnitudes for the same representative joints show attenuation and redistribution.
\textbf{e,} Cumulative reward and fixed-condition evaluation reward.}
\label{fig:go2_case}
\vspace{-0.9em}
\end{figure*}

\subsection{Experimental Protocol}
Robust control is evaluated under joint-level health degradation.
For each controllable joint \(j\), an undamaged actuator has health value \(h_{t,j}=1.0\).
Degraded joints take health values from \(\{0.7,0.5,0.3,0.0\}\), where smaller values indicate more severe loss of control authority and \(0.0\) corresponds to a fully disabled action channel.
Unless otherwise specified, the degradation mask and health values remain fixed within an episode.
HealthNet uses a history length of \(K=10\), taking the most recent observation--action history to estimate the current joint-health vector, which is then concatenated with the observation as policy input.
All reported results are averaged over three random seeds.

\textbf{Platforms and tasks.}
The evaluation spans four robot platforms: (i) DMControl Walker~\cite{tassa2018dmcontrol}, (ii) Unitree Go2~\cite{unitreeGo2}, (iii) Cassie~\cite{gong2019cassie}, and (iv) a Panda arm~\cite{gaz2019panda} in ManiSkill~\cite{mu2021maniskill,gu2023maniskill2}.
For Walker, we evaluate four planar locomotion tasks: Walk, Run, Walk-Backwards, and Run-Backwards.
Walker has six controllable action joints.
All degradation combinations with \(d\) from \(0\) to \(3\) degraded joints are evaluated, yielding \(1{,}545\) health conditions per task.
Across the four tasks, this gives \(6{,}180\) total health conditions; when reporting robustness under degraded conditions, we exclude the four mask-0 healthy cases, resulting in \(6{,}176\) degraded Walker conditions.

For Unitree Go2, we evaluate flat- and rough-terrain velocity tracking in Isaac Lab~\cite{mittal2025isaaclab}.
Go2 has twelve action-target joints, and degradation is applied by scaling the target-action offsets with the corresponding health values.
Degradation masks with \(d=1\)--\(8\) degraded joints are evaluated using a fixed sampling grid: \(48\) one-joint degradation conditions and \(256\) conditions for each \(d=2,\ldots,8\), giving \(1{,}840\) health conditions per Go2 task.

For Cassie, we adopt the same flat- and rough-terrain velocity-tracking protocol.
Cassie provides a bipedal counterpart in which joint degradation directly challenges leg coordination and balance.
As with Go2, degradation is applied at the action-target level over twelve controlled joints, and each Cassie task is evaluated on \(1{,}840\) health conditions covering \(d=1\)--\(8\) degraded joints.

For manipulation, we evaluate Panda-based ManiSkill tasks, including StackCube, OpenCabinetDrawer, and PickSingleYCB~\cite{calli2015ycb}.
The Panda arm has seven controllable arm joints, with the gripper left unmasked.
Degradation masks with \(d=1\)--\(4\) degraded arm joints are evaluated; for each \(d\), all joint masks are enumerated and assigned a shared degraded-joint health value from \(\{0.7,0.5,0.3,0.0\}\), yielding \(392\) health conditions per manipulation task.
These tasks assess health-conditioned control in contact-rich manipulation beyond legged locomotion.

\textbf{Evaluation scale.}
For locomotion, the benchmark contains \(6{,}180\) Walker health conditions and \(7{,}360\) Isaac Lab Go2/Cassie health conditions, for a total of \(13{,}540\) unique locomotion health conditions.
The evaluation counts below cover the nominal baseline and the health-conditioned policy.
Both are evaluated on every condition, yielding \(27{,}080\) policy-condition evaluations before seeds and \(81{,}240\) evaluations across three seeds.
Panda manipulation contributes \(1{,}176\) health conditions across three tasks, bringing the full benchmark to \(14{,}716\) unique health conditions and \(29{,}432\) policy-condition evaluations before seed averaging.

\textbf{Baselines.}
Two baselines are compared.
The nominal controller is trained without degradation-aware conditioning and is evaluated under the same degraded test conditions.
The degradation-trained no-health controller uses the same degradation distribution, action degradation mechanism, optimizer, network size, reinforcement learning algorithm, and training budget as the proposed method, but removes HealthNet and the health-conditioned policy input.
This baseline tests whether exposure to randomized degradation alone suffices without an explicit per-joint health representation.

\textbf{Training algorithms and HealthNet integration.}
The underlying reinforcement learning algorithm for each benchmark is kept unchanged, and HealthNet is integrated as a health-estimation and policy-conditioning module.
For the DMControl Walker tasks, including Walk, Run, Walk-Backwards, and Run-Backwards, the nominal baseline is trained with TD-MPC2~\cite{hansen2024tdmpc2}.
The proposed method uses the same TD-MPC2 training pipeline, augmented with HealthNet to estimate the joint-health vector from the recent observation--action history.
The predicted health vector is provided as part of the policy input and is trained with the auxiliary health prediction loss in~\eqref{eq:health_loss}.

For Isaac Lab locomotion, the nominal baseline uses PPO with the RSL-RL implementation~\cite{schulman2017ppo,schwarke2025rslrl}.
The proposed method adopts the same PPO training pipeline and augments the policy observation with the HealthNet prediction.
For Go2, HealthNet predicts the health values of the 12 action joints; for Cassie, it predicts the health values of the controlled action joints.
In both cases, the estimated health vector is concatenated with the policy observation, while the true sampled health is used only during simulation training for action degradation and auxiliary supervision.
At test time, the policy receives only the predicted health and has no access to the ground-truth health vector.

\textbf{Metrics.}
Locomotion reports episode return; manipulation reports success rate, with return as a secondary graded metric.

\subsection{Results}

Table~\ref{tab:main_results} summarizes performance under joint degradation. The nominal controller degrades substantially, and the degradation-trained no-health baseline remains consistently below the health-conditioned policy despite experiencing the same degradation distribution. This indicates that randomized degradation exposure alone is insufficient for structured compensation.

The health-conditioned policy improves performance on all evaluated tasks. Walker returns increase by \(22.5\%\)--\(72.4\%\), while Isaac Lab gains reach \(191.6\%\) on Go2 Flat and \(250.1\%\) on Cassie Flat. On Panda-based ManiSkill tasks, success rates improve by \(5.9\)--\(15.2\) percentage points. In addition, episode returns increase by \(11.2\%\), \(13.5\%\), and \(5.6\%\) on StackCube, OpenCabinetDrawer, and PickSingleYCB, respectively.

\begin{table}[!t]
\centering
\caption{MANISKILL EPISODE RETURNS UNDER JOINT DEGRADATION. GAINS ARE COMPUTED RELATIVE TO THE NOMINAL BASELINE.}
\label{tab:maniskill_return}
\small
\renewcommand{\arraystretch}{1.10}
\begin{tabular*}{\columnwidth}{@{\extracolsep{\fill}}lccc@{}}
\toprule
\textbf{Task} & \textbf{Nominal} & \textbf{Ours} & \textbf{$\Delta_{\mathrm{Nominal}}$} \\
\midrule
StackCube & 21.2705 & 23.6560 & +11.2\% \\
OpenCabinetDrawer & 60.1089 & 68.2218 & +13.5\% \\
PickSingleYCB & 14.5077 & 15.3222 & +5.6\% \\
\bottomrule
\end{tabular*}
\vspace{-0.6em}
\end{table}

Fig.~\ref{fig:go2_case} provides a fixed-degradation case study on Go2. Under the same five-joint degradation condition, the baseline loses stability, while the proposed policy maintains forward locomotion. For visualization, health and action statistics are plotted for three representative degraded joints, showing attenuation on degraded joints together with redistribution across the remaining actuators.

\subsection{Robustness Across Morphologies and Degradation Severity}
\begin{figure}[!t]
\centering
\includegraphics[width=\columnwidth]{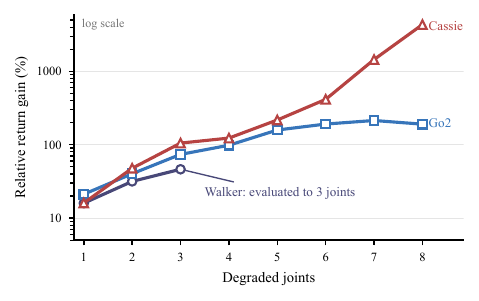}
\caption{Relative return gain over the nominal baseline by degraded-joint count. Walker is evaluated only for \(d=1\)--\(3\); its curve ends at the last evaluated point. Go2 and Cassie cover \(d=1\)--\(8\). The vertical axis is logarithmic.}
\label{fig:severity}
\vspace{-0.8em}
\end{figure}

We next analyze performance as degradation severity increases. Table~\ref{tab:severity} summarizes gains for low, medium, and high degradation; the platform-specific joint-count ranges are defined below the table. These groups span all sampled health values within their respective joint-count ranges and do not imply equal physical severity across platforms. Fig.~\ref{fig:severity} shows that relative return gain generally increases with the number of degraded joints across locomotion platforms. The Walker curve ends at \(d=3\), its maximum evaluated degraded-joint count; Go2 and Cassie are evaluated through \(d=8\).

\begin{table}[!t]
\centering
\caption{Gain by degradation level relative to the nominal baseline. Locomotion: relative return gain (\%); Panda: success-rate gain (pp, percentage points).}
\label{tab:severity}
\small
\setlength{\tabcolsep}{5pt}
\renewcommand{\arraystretch}{1.10}
\begin{tabular*}{\columnwidth}{@{\extracolsep{\fill}}lccc@{}}
\toprule
\textbf{Platform} & \textbf{Low} & \textbf{Medium} & \textbf{High} \\
\midrule
Walker & +29.8\% & +46.3\% & +62.1\% \\
Go2 & +36.8\% & +103.8\% & +198.3\% \\
Cassie & +41.6\% & +139.1\% & +928.0\% \\
\midrule
Panda & +12.3 pp & +11.6 pp & +10.8 pp \\
\bottomrule
\end{tabular*}
\par\vspace{3pt}
\begin{minipage}{\columnwidth}
\footnotesize\raggedright
\textit{Degraded-joint counts (Low / Medium / High):}
Walker, 1 / 2 / 3; Go2 and Cassie, 1--2 / 3--5 / 6--8;
Panda, 1 / 2--3 / 4.
\end{minipage}
\end{table}

Table~\ref{tab:severity} shows increasing locomotion gains from low to high degradation. The largest relative gain, \(928.0\%\) on Cassie, corresponds to a return increase from \(3.94\) to \(40.50\), reflecting the low nominal return under severe degradation. Panda success-rate gains remain positive across groups at \(10.8\)--\(12.3\) percentage points.

The improvement extends beyond selected favorable masks: the proposed method outperforms the nominal baseline in \(92.3\%\) of Walker, \(94.3\%\) of Go2, \(84.2\%\) of Cassie, and \(81.8\%\) of Panda conditions. These experiments support the health-conditioned principle across different morphologies; each platform uses a separately trained policy.

\subsection{Ablation Analysis}
Table~\ref{tab:ablation} evaluates key components.

\begin{table}[!t]
\centering
\caption{Ablation of key components.}
\label{tab:ablation}
\small
\setlength{\tabcolsep}{6pt}
\renewcommand{\arraystretch}{1.10}
\begin{tabular*}{0.9\columnwidth}{@{\extracolsep{\fill}}lcc@{}}
\toprule
\textbf{Variant} & \textbf{Policy Input} & \textbf{Perf.} \\
\midrule
Full & \(o_t,\hat h_t\) & 100.0\% \\
No health cond. & \(o_t\) & 74.6\% \\
Scalar health & \(o_t,\hat h^{\rm scalar}_t\) & 82.3\% \\
No degradation & \(o_t,\hat h_t\) & 70.1\% \\
No HealthNet & \(o_t\) & 65.4\% \\
\bottomrule
\end{tabular*}
\end{table}
Removing health conditioning reduces performance to \(74.6\%\), demonstrating the benefit of providing predicted health to the policy. Scalar health reaches \(82.3\%\), indicating that global severity cannot capture heterogeneous joint failures. Removing the action degradation gate reduces performance to \(70.1\%\), and removing HealthNet gives the largest drop to \(65.4\%\). These findings support the combined use of degradation-aware training, per-joint health estimation, and health-conditioned policy execution.

\section{Conclusion}
In this paper, a joint-health-aware reinforcement learning framework was presented for robust robot control under simulated actuator-authority degradation. The method models residual actuation capability through a continuous per-joint health vector. A HealthNet model infers this vector from recent observation--action history, enabling the policy to condition its decisions on predicted health and dynamically redistribute control effort accordingly. Experiments across Walker, Go2, Cassie, and Panda-based manipulation tasks show consistent robustness improvements under diverse degradation masks and severity levels. Future work will extend the framework to real hardware and encompass richer failure modes such as time-varying degradation, joint locking, response delays, and combined sensing--actuation faults.

\begingroup
\renewcommand{\footnotesize}{\fontsize{8}{9.2}\selectfont}
\providecommand{\BIBdecl}{}
\renewcommand{\BIBdecl}{\setlength{\itemsep}{0pt}\setlength{\parsep}{0pt}\setlength{\parskip}{0pt}}
\setlength{\itemsep}{0pt}
\setlength{\parskip}{0pt}
\bibliographystyle{IEEEtran}
\bibliography{references}
\endgroup

\end{document}